# QuASAR: Toward an LLM-Assisted Framework for Screening, Prototyping, and Assessing Quantum-Advantage Opportunities in Scientific Computing

Mark L. Carroll [1a], Jordan A. Caraballo-Vega[1], Brian P. Powell [1], John L. Schnase [1,2]

[1] NASA Goddard Space Flight Center
[2] Adnet System, Inc.

[a] Corresponding author Mark.Carroll@nasa.gov

## Abstract

Identifying scientific computations that could benefit from quantum algorithms requires both domain knowledge and fluency in quantum algorithm design, expertise that rarely resides in one person. We report results from QuASAR, an experimental, two-agent, LLM-driven pipeline intended to bridge this gap. One agent screens scientific literature for computational kernels with a plausible quantum-algorithmic analogue and proposes candidates, each backed by citations. A second agent independently verifies those citations. For candidates that survive verification, it reconstructs the source's computational structure, formulates a quantum mapping, builds and executes a prototype, and benchmarks it against classical baselines. We applied this pipeline to ten candidate computations drawn from the astrophysics literature. Nine candidates were carried through to a small-scale, benchmarked prototype. We report three in detail, chosen as the clearest examples with the least redundancy across the set. They span combinatorial optimization, stochastic simulation, and Bayesian inference. The pipeline attempted to supply quantum-design expertise a domain scientist might lack, reconstructing source algorithms where one existed and building original, disclosed mappings where none did. None of the three case studies achieved quantum advantage at the tested scales, and their astrophysical framing has not yet undergone expert validation. What the experiment does show is that the pipeline can turn a published idea into a runnable prototype, which a scientist can then inspect, test against classical methods, and decide whether to pursue. QuASAR is an exploratory framework, offered to the community for evaluation, critique, and further development, with an approach that could broaden the use of quantum computing in scientific research and extend to computational improvements more generally.

## 1. Introduction

Quantum computers promise theoretical asymptotic speedups for a handful of computational primitives, such as sampling and mean estimation, linear systems, search, and simulation of quantum dynamics [1]. Turning that promise into working prototypes for a specific scientific computation, however, remains slow and easy to get wrong. While astrophysicists, climate modelers, computational biologists, and other domain scientists know their computational bottlenecks, they rarely have training in quantum algorithm design, circuit construction, or the practicalities of encoding classical data onto qubits. Quantum-algorithms researchers, on the other hand, have that training, but typically lack the domain context to judge whether a proposed quantum mapping preserves the scientific content of the original problem.

Building a genuine quantum prototype no longer requires access to quantum hardware. Accessible SDKs such as Qiskit [2], and GPU-accelerated simulators such as NVIDIA's CUDA-Q and cuQuantum, let researchers write quantum programs and run them in simulation on classical infrastructure [3]. What remains scarce is not access but the fluency to use it. This asymmetry is not unique to astrophysics. A recent survey of practicing quantum software developers found that formal training bridging quantum physics and software engineering was rare even among a highly credentialed sample, and one practitioner described someone fluent in both as almost impossible to find [4]. A recent review of machine learning for quantum computing identifies the same divide and expects LLM-based coding assistants to help close it [3].

QuASAR (Quantum Advantage Screening, Assessment, and Research) is an experimental framework designed to explore one way of addressing this gap. QuASAR comprises two pipelined LLM-based agents. Agent A screens scientific literature for computational kernels that might be amenable to quantum acceleration and proposes candidates. Agent B independently resolves Agent A's citations and verifies the proposed candidates. For valid candidates, Agent B rebuilds the source's computational structure, formulates a quantum mapping, then constructs and benchmarks a small-scale prototype in simulation.

We tested this process on ten candidate computations from one screening pass over the astrophysics literature curated in the Astrophysics Data System (ADS) and Science Explorer (SciX). Verification caught a citation problem in two candidates. One had no astrophysical grounding in any source and was dropped as a candidate. The other's citation was corrected and disclosed, and the candidate was carried through like the rest. Nine candidates in total were carried through reconstruction, prototyping, and benchmarking. This note reports three in detail. They span Bayesian inference [5], combinatorial optimization [6], and stochastic simulation [7, 8], chosen as the clearest examples with the least redundancy across the full set. The other six are documented to the same standard in the accompanying archive (Section 4.4).

The work described here is experimental and limited. We do not claim that the three case studies are validated astrophysical applications, novel quantum algorithms, or demonstrations of quantum advantage. Astrophysics served as a technically rich proving ground for the pipeline, not as this paper's subject. The question for us has been whether an LLM-assisted workflow can generate candidate quantum mappings that are concrete enough to inspect, run, benchmark, and reject or refine. We release the resulting code and reconstruction records so that domain scientists and quantum-computing researchers can evaluate that question directly and find errors we have missed. QuASAR's underlying workflow may also apply to other computational-improvement searches in published documents or scientific code sets. A framework such as QuASAR could screen for GPU acceleration, faster classical subroutines, or newly published numerical methods.

Section 2 describes the two-agent workflow, verification protocol, reconstruction rules, prototyping, benchmarking, and evidence tiers. Section 3 reports the ten-candidate triage and the three completed case studies. Section 4 discusses what the experiment supports, its limitations, and reproducibility.

# 2. Materials and Methods

## 2.1 A two-agent pipeline

Large language model (LLM) agents can generate hypotheses, write and execute code, and automate parts of scientific workflows [9]. Comparable tool-augmented agents exist for chemistry [10], for translating legacy scientific code, including multi-agent designs that outperform single-pass translation [11, 12], and for automating hypothesis generation and experimental validation in drug discovery [13]. QuASAR tests whether two such agents, working in sequence, can screen scientific literature for quantum-computing opportunities and build something concrete enough to check.

Agent A screens scientific material for computational kernels with a plausible quantum-algorithmic analogue and proposes candidates, each supported by citations to primary sources. Agent B carries out two further steps. Step one is verification. Agent B independently resolves Agent A's citations and checks whether each cited source supports the application Agent A attached to it. Step two is reconstruction, and it applies only to candidates that pass step one. Agent B rebuilds the source's computational structure from the primary literature, formulates a quantum mapping, and constructs, executes, and benchmarks a prototype. Agent A was implemented using OpenAI's GPT-5.5 [14]; Agent B was implemented using Anthropic's Claude Sonnet 5 [15]. This split was deliberate, not incidental. Section 2.3 explains the reasoning behind it. Section 4.3 discusses what the design does and does not guarantee.

## 2.2 Agent A: quantum-opportunity screening

Agent A was given a broad astrophysics research context and asked to identify computationally demanding problems that might admit a quantum-algorithmic speedup, such as Monte Carlo estimators, differential-equation solvers, combinatorial optimizers, Bayesian inference routines. Each proposal needed citations to both the classical astrophysics problem and a corresponding quantum-algorithms paper. A single screening pass produced ten candidate problems.

## 2.3 Agent B, step 1: source resolution and claim verification

LLMs are generally poor evaluators of their own output. The same models that catch and correct errors in externally sourced content often miss identical errors in their own reasoning traces, an asymmetry tied to how a claim is labeled rather than a hard capability limit [16-18]. A recent survey of verification mechanisms for LLM-based scientific agents ranks self-correction, a model checking its own output, as the weakest of four major strategies, behind multi-agent critique, human-in-the-loop oversight, and tool-based validation [19]. Agent B's verification step used a different model family than Agent A's screening step for this reason.

For each candidate, Agent B independently resolved every citation Agent A had used by using the web search and fetch tools in its Claude environment to look up its arXiv identifier and DOI, retrieve the arXiv record and, where available, the full text, and compare title, authors, venue, and year against Agent A's citation. It read each primary source in full wherever feasible. It checked specific technical claims, stated equations, complexity results, and whether the source's stated application matched the astrophysical framing Agent A had attached to it. A candidate passed if its citations were bibliographically correct and topically on point (Section 3.1). To test what Agent B does when there is nothing to find, Agent B was given three real, actively practiced astrophysical computations with no Agent A proposal behind them: catalog cross-matching, gravitational N-body dynamics via tree codes, and CCD image detrending. Each was chosen because it was expected, going in, to have no viable quantum accelerator. In these three null cases, Agent B searched the literature directly and reached its own judgment, with no Agent A candidate proposal to react to (Section 3.4).

## 2.4 Agent B, step 2: reconstruction, quantum formulation, and prototyping

For each candidate that passed verification, Agent B reconstructed the primary source's computational structure independently, working from the primary literature rather than Agent A's summary. It then formulated a quantum mapping for that structure and classified how faithfully a reduced-scale instance of that mapping could reproduce the source's structure, using four categories. *Exact* means the reduced-scale instance reproduces the source's mathematical structure without simplification. *Approximate* means a specific, disclosed simplification stands in for part of the source's method. *Surrogate* means the reduced-scale instance captures the problem's qualitative structure using synthetic or reduced data rather than the source's own dataset. *Hybrid* means the prototype combines quantum and classical subroutines as the source itself specifies. Any simplification was logged in a disclosed-simplification ledger stating what was substituted and why. Agent B then built an executable prototype in Python using Qiskit [2], ran it in simulation on classical hardware (rather than a physical quantum processor) and recorded the results.

## 2.5 Simulation and classical benchmarking

Each prototype was benchmarked against two classical references: an exact or brute-force solver, checking correctness at small problem sizes, and a strong classical heuristic or standard numerical method, checking competitiveness. This two-tier design is intended to avoid a well-documented failure mode. Comparing a quantum prototype only against a weak classical baseline can mask or fake an apparent speedup [20].

All Python and Qiskit code was re-run on local machines, outside the Claude environment in which Agent B originally developed and executed it. These reruns provided a check on whether reported outputs could be reproduced outside that environment, not an artifact of it.

### 2.6 Quantum advantage assessment

Each case study was assigned an evidence tier for its quantum-advantage claim. *Demonstrated* means the prototype itself shows a measured advantage at the tested scale. *Extrapolated* means the prototype's measured scaling behavior implies an advantage at larger problem sizes not yet observed directly. *Theoretical* means the case for advantage rests on the source literature's complexity results, not on anything measured here.

## 3. Results

### 3.1 Candidate triage across ten problems

Table 1 summarizes the outcome of Agent B's verification step across all ten candidates from the screening pass in Section 2.2. Verification found no astrophysical grounding for one candidate in any cited source; it was dropped as an astrophysics prototype, though a non-astrophysical technical demonstration of the same underlying method was built for completeness. Verification found and disclosed a correctable citation error for a second candidate, which was then carried through like the rest. The remaining eight passed cleanly. Three of the nine astrophysics prototypes become the detailed case studies below. The other six were built and benchmarked to the same standard but are not narrated in detail here for the reason stated in the table.

**Table 1.** Summary of the Agent B verification step.

| # | Problem | Outcome | Primary source(s) | Reason / status |
|---|---|---|---|---|
| 1 | Vlasov simulation of large-scale structure with massive neutrinos | Prototyped, not detailed | Miyamoto et al. 2024 [21] | Read in full; precise match. Built and benchmarked; depends on qRAM and a linearized-physics simplification, a less central example than the three detailed here. |
| 2 | Schrödinger-Poisson surrogate dark-matter simulation | Prototyped, not detailed | Cappelli et al. 2024 [22] | Read in full and on-topic. Built and benchmarked; a clean, well-supported candidate not among the three chosen for detailed reporting. |
| 3 | Gravitational-wave Bayesian parameter estimation | Case study 1 | Escrig et al. 2025 [5] | Read in full. Already both astrophysics and a quantum algorithm; the most direct match found. |
| 4 | Pulsar timing array inference for stochastic GW backgrounds and anisotropy | Corrected, prototyped | Hazboun et al. 2020 [23] (verified) | The second citation, given as a 2025 paper, resolved to a mismatched source about a different detector. Agent B substituted the correct match, Ding & Tian 2024 [24], disclosed the substitution, and carried the candidate through to a prototype. Unlike every other prototyped candidate, no source proposes the quantum mapping itself. |

| # | Problem | Outcome | Primary source(s) | Reason / status |
|---|---|---|---|---|
| 5 | Time-domain transient discovery in X-ray survey catalogs | Prototyped, not detailed | Kawamuro et al. 2025 [25] | Primary citation read in full and directly astronomical; three supporting citations verified only at abstract level. Built and benchmarked; less central than the three detailed here. |
| 6 | Survey scheduling as combinatorial optimization | Case study 2 | Chancellor 2019 [6] | All five citations verified exactly. The astrophysical application is Agent B's own disclosed analogy. |
| 7 | Poisson/elliptic solver for astrophysical mapmaking and lensing mass-mapping | Failed verification | Choi & Ryu 2025 [26]; Liu et al. 2025 [27] | Both citations resolved and read in full. Neither addresses astrophysics; the framing was invented. A non-astrophysical technical demonstration of the underlying Poisson-solver method was built for completeness and is included in the archive, but is not counted as an astrophysics prototype. |
| 8 | Dissipative PDE evolution and transport via LCHS | Prototyped, not detailed | Novikau & Joseph 2025 [28] | Citation verified, though only the paper's introductory sections were read in full. A simpler generic construction was substituted for the paper's own more advanced QSP-based circuit, which was not reconstructed. |
| 9 | Monte Carlo-dominated astrophysical inference, including radiative transfer | Prototyped, not detailed | Montanaro 2015 [8]; Rebentrost et al. 2018 [29]; Destainville et al. 2010 [30] | All three checked out bibliographically, but only at abstract level. Built and benchmarked; the same amplitude-estimation kernel is covered more thoroughly, with full-text verification, in Problem 10. |
| 10 | Stochastic differential equation models of astrophysical time-domain variability | Case study 3 | An et al. 2021 [7], building on Montanaro 2015 [8] | The most exhaustively verified claim set of any candidate, applied to a genuinely astrophysical technique. |

## 3.2 Case studies

***Problem 3*** – Escrig et al. describe qBIRD, a hybrid quantum algorithm for gravitational-wave Bayesian parameter estimation that combines a quantum walk with a classical renormalization-and-downsampling procedure [5]. The quantum walk moves over a discretized grid of parameter values. A coin register selects a direction, an ancilla qubit encodes the Metropolis-Hastings acceptance probability [31] as a rotation angle, and the position register shifts only when that ancilla measures an accept. Agent B reconstructed three of the walk operator's five sub-operators (the coin, acceptance, and shift steps) and implemented them as an executable Qiskit prototype, using a small number of qubits to represent a single parameter, standing in for a quantity such as chirp mass, discretized over a small grid. It did not reproduce the walk operator's remaining sign-flip and reflection sub-operators or the paper's renormalization-and-downsampling outer loop, and disclosed these omissions rather than presenting the core-only reconstruction as complete. This case study was classified as Hybrid and Approximate. It reproduces the paper's central sampling mechanism but omits a stated component of the full algorithm. qBIRD is itself an astrophysics paper describing a quantum algorithm, so this candidate needed no analogy between a generic method and an astrophysical target. The two were already the same paper. The full analysis for this and the other two case studies is documented in the accompanying QuASAR output package (Section 4.4).

***Problem 6*** – Chancellor proposes a domain-wall encoding of discrete variables for quantum annealing and the Quantum Approximate Optimization Algorithm (QAOA) [32], tested on synthetic scheduling and graph-colouring problems rather than astronomical data [6]. The encoding represents a discrete variable's chosen value by the position of a single transition point, a domain wall, along a short chain of qubits, using one fewer qubit per variable than a one-hot encoding. QAOA then alternates a cost unitary and a mixer unitary across that register, with a classical optimizer tuning both to concentrate measurement probability on low-cost, valid assignments. Agent B reused this encoding for a reduced-scale astronomical survey-scheduling problem formulated as a Quadratic Unconstrained Binary Optimization (QUBO), an application the source paper does not address, disclosed as Agent B's own construction rather than attributed to the source. Two disclosed simplifications were used: a standard QAOA mixer in place of the constrained, subspace-preserving mixer more consistent with domain-wall variables, and a brute-force diagonal construction of the QUBO's conflict term in place of the compact two-body Ising form Chancellor's method would produce at scale. An initial QAOA run converged to a local optimum; a multi-restart procedure fixed it and is reported here as a negative result, not omitted. The prototype was benchmarked against simulated annealing on the same QUBO instance; simulated annealing solved it essentially instantly, while the QAOA circuit needed additional restarts and circuit depth before it reliably measured the correct answer. This case study was classified as Approximate.

***Problem 10*** – An et al., building on a general quantum Monte Carlo speedup due to Montanaro [8], describe a quantum-accelerated multilevel Monte Carlo (MLMC) [33] method for stochastic differential equations [7]. MLMC combines a coarse and a fine discretization of the same random path, so that only the difference between the two levels needs a costly, high-precision estimate, reducing the total number of samples needed for a target accuracy compared with a single-level estimator. Agent B applied this method to a reduced-scale Ornstein-Uhlenbeck process, a technique genuinely used elsewhere in astrophysics to model stochastic time-domain variability. The specific reduced-scale instance is Agent B's own construction, not a fit to real observational data. Agent B reconstructed the source's two-level coupling structure, generating fine-level Brownian increments first and computing the coarse-level increment as their sum over the same path, preserving the correlation between levels that gives MLMC its variance reduction. The one disclosed simplification was a Grover-power-sweep estimator, which infers a mean from measurement statistics gathered across repeated applications of a Grover-type operator [34] at increasing power [35], in place of the source's formal amplitude-estimation subroutine, which uses phase estimation for a provable quadratic speedup. At the reduced problem sizes tested, the quantum prototype's estimate was more accurate than either classical baseline at a comparable sample budget; because the demonstration used only two discretization levels at first discretization order, this does not demonstrate the source's claimed asymptotic advantage, which requires more levels and a higher-order scheme. This case study was classified as Approximate.

## 3.3 Summary across case studies

None of the three case studies reached the Demonstrated tier at the sizes tested. The gravitational-wave case study (Problem 3) rests on Theoretical evidence, the source paper's complexity claims, not a comparison measured here. The scheduling case study (Problem 6) was directly measured, and the classical baseline won: simulated annealing solved the instance essentially instantly against a QAOA circuit that needed extra depth and restarts to reliably find the right answer. The multilevel Monte Carlo case study (Problem 10) was also directly measured; there, the quantum prototype's estimate was more accurate than the classical baselines at a comparable sample budget, though only two discretization levels were tested, short of the regime the source's asymptotic advantage requires. None of the three offers even an Extrapolated case for advantage. What the three case studies do show is that Agent B turned a published method into a runnable circuit, checked it against an exact classical solver, compared it with a strong classical method, and recorded every departure from the source. Detailed benchmark results, circuit specifications, and the full reconstruction record for each case study are available as a separate repository (Section 4.4).

### 3.4 When there is nothing to find

Section 2.3 describes the rationale for this check. Agent B was given three real, actively practiced astrophysical computations with no Agent A proposal behind them: catalog cross-matching, gravitational N-body dynamics via tree codes, and CCD image detrending. In each case, Agent B found quantum-computing papers on related problems, but none of them yielded a viable quantum method for the specific computation being tested. The archive documents why, case by case (Section 4.4). No quantum prototype was built. The classical baseline was implemented and benchmarked instead and checked analytically against the quantum literature's own reported costs. All three concluded no plausible quantum advantage. Across all three tests, Agent B reported that there was nothing to find rather than manufacturing a positive result. This does not establish that every claim Agent B makes about a genuine candidate is accurate; that is what the full-text verification in Section 2.3 is for. However, it is consistent with the narrower and more basic point that Agent B did not fabricate a match when none existed in these tests.

## 4. Discussion

### 4.1 What this paper does and does not claim

This is early work and does not at this point represent a validated astrophysical application, nor does it introduce a novel quantum algorithm or a demonstrated quantum advantage. With QuASAR, we have provided an example of a two-agent, LLM-assisted screen-verify-build pipeline, which, when applied to ten literature-derived problems, correctly identified citation issues in two of them, dropping one and correcting the other, and produced nine benchmarked astrophysics prototypes, each with its simplifications disclosed. Three case studies are reported here in full detail; the remaining six are documented in the accompanying archive (Section 4.4).

A separate check (Section 3.4) showed that, in three tests where there was nothing to find, Agent B reported nothing rather than manufacturing a result. Given three real computations with no Agent A proposal behind them at all, Agent B still reached, and defended in writing, three negative results that matched our expectations going in. A related, more informal test made the same point more vividly. During pipeline development, Agent A's source document was seeded with a synopsis of Harper Lee's *To Kill a Mockingbird*. Agent B's response identified the item as having nothing for the pipeline to act on and dropped it before any formal analysis began.

Collectively, the three case studies also illustrate what closing the gap between domain expertise and quantum fluency might look like in practice. In the gravitational-wave case (Problem 3), a genuine quantum algorithm already existed in the literature, and Agent B's contribution was to reconstruct its core and disclose what it left out. In the scheduling and multilevel Monte Carlo cases (Problems 6 and 10), no cited source proposed an astrophysical application, so Agent B attempted to supply the missing quantum-design expertise, grounding a new mapping in a verified generic method rather than inventing one from nothing. What distinguishes both roles from an ordinary literature summary is that the pipeline discloses, case by case, which one it is playing.

### 4.2 The shape of gap-closing across the full set

Agent A supplies the domain framing; Agent B supplies the quantum fluency. Verification is meant to keep that fluency honest by checking Agent A's framing against the primary sources before reconstruction proceeds. Table 1 records how this plays out across all ten candidates. Read alongside the case studies, it shows that closing the gap between domain expertise and quantum expertise takes different forms depending on what verification finds.

In some cases, the literature already contained a genuine quantum algorithm, so Agent B's task was to verify and reconstruct it faithfully. qBIRD (Problem 3) is the clearest example. Source and target were already the same paper. Problem 1 follows the same pattern. Agent B's reconstruction of Miyamoto et al.'s algorithm was judged a "precise match," preserving the paper's own qRAM dependency and linearized-physics simplification rather than introducing new ones. Problem 10 belongs in this category as well, with "the most exhaustively verified claim set of any candidate, applied to a genuinely astrophysical technique."

In other cases, verification found no bridge in the literature at all. Agent B therefore had to construct a mapping that no source proposed. For Problem 6, the citations were "verified exactly," but the astrophysical application is explicitly identified as "Agent B's own disclosed analogy." Problem 4 goes further. Unlike every other prototyped candidate, "no source proposes the quantum mapping itself." The mapping is therefore entirely Agent B's construction.

A third pattern lies between these two. When verification could confirm a source only partially, Agent B's fluency took the form of a generic construction standing in for a method it could not fully verify. In Problem 8, "a simpler generic construction" was substituted for the paper's more advanced QSP-based circuit, which was not reconstructed. Problem 9 presents a similar limitation at the level of evidence. All three citations were verified bibliographically, but only "at abstract level," a thinner foundation than the full-text verification supporting the same estimation kernel in Problem 10.

These are not three separate outcomes so much as one mechanism responding to three states of the literature: a real bridge to reconstruct, no bridge and therefore one to build, or a partial bridge whose gaps must be filled and explicitly flagged. None of the ten candidates ends in a deployed solution because we were simply testing for mechanisms not deployable solutions. What each produces instead is a documented record of what closing the domain-quantum gap required in that particular case.

That record points in two directions at once. For a domain scientist, a verified reconstruction, or a disclosed, literature-grounded analogy, provides something concrete to inspect, question, and evaluate rather than simply take on faith. This may lower the threshold for engaging with quantum computing in the first place. For a quantum-algorithms researcher, the same record suggests where a technique developed for one purpose may apply in a domain its original authors never considered. Chancellor's domain-wall encoding, for example, was developed for generic scheduling problems. Here it is mapped onto astronomical survey scheduling.

Seen this way, the case studies are more than evidence about the performance of a single pipeline. They are examples of the bridge-building that may be needed on both sides of the domain-quantum divide.

### 4.3 Limitations

Several limitations bound what this work supports. Every case study operates at reduced problem sizes chosen for tractable classical simulation, not the scale of any real dataset. Every prototype ran on a classical simulator, not physical quantum hardware; none of the reported circuit behavior reflects real-device noise, decoherence, or qubit connectivity, any of which could change results at any scale. None of the reported comparisons should be read as evidence about performance at realistic astrophysical scale. The Exact / Approximate / Surrogate / Hybrid classification depends on Agent B's own judgment about what counts as a faithful reconstruction, a judgment not yet subjected to independent human review. The verification step (Section 2.3) checks whether a citation exists and supports its stated application. It does not verify that a proposed quantum mapping is a good use of quantum resources for the underlying problem, and it may miss a citation-and-application mismatch that a domain expert would catch but our protocol was not designed to test for. Agent A and Agent B were implemented using different model families, OpenAI's GPT for screening and Anthropic's Claude for verification and reconstruction, a stronger independence guarantee than prompting a single model to play both roles, but not a clean one. Both are large language models trained on overlapping web and scientific-literature corpora. Residual correlation between Agent A's and Agent B's judgments cannot be ruled out.

### 4.4 Reproducibility

The complete QuASAR output package accompanies this preprint as a permanent Zenodo archive [36]. It includes Agent A's literature-scan output, the prompt files used to instruct Agent A and Agent B, Agent B's source-verification summary for all ten candidates, the nine astrophysics-candidate repositories, plus a tenth, non-astrophysical technical demonstration built for the candidate whose citations supported no astrophysical claim (Problem 7), and separate folders for the null tests and the local-machine confirmation runs.

## Conclusion

QuASAR is a two-agent, LLM-driven pipeline that screens scientific literature for quantum-advantage opportunities, verifies proposed candidates, reconstructs a viable candidate's computational structure, formulates a quantum mapping, and builds, executes, and benchmarks a prototype against classical baselines. Applied to ten candidate computations drawn from the astrophysics literature, it carried nine through to a small-scale, benchmarked prototype; this note reports three in full. Across those three, closing the gap between scientific domain expertise and quantum computing expertise took different forms. Agent B reconstructed the core of a genuine source algorithm in one case and built an original, disclosed mapping in the others, with Agent A's domain framing and Agent B's quantum fluency checked against each other before any prototype was built. QuASAR illustrates a process, not a finished system, and we offer it to the community for examination, critique, and extension. The underlying discipline is not specific to quantum computing: propose, verify, reconstruct, build, and benchmark against a real baseline could apply as readily to GPU acceleration, faster classical algorithms, and other computational-improvement opportunities.

## Acknowledgments

The authors thank Dr. Jeremy D. Schnittman of the Astrophysics Science Division at NASA Goddard Space Flight Center for helpful comments and feedback on the manuscript.

## LLM Assistance Disclosure

The authors used Claude Sonnet 5 [15] as an LLM writing assistant to help with drafting, editing, and refining the manuscript text. All LLM-assisted content was reviewed and edited by the authors, who independently verified the accuracy of the final text against source materials. The authors take full responsibility for the accuracy and integrity of the published work.